\documentclass[11pt]{article}

\usepackage[preprint]{acl}

\usepackage{times}
\usepackage{latexsym}

\usepackage[T1]{fontenc}

\usepackage[utf8]{inputenc}

\usepackage{microtype}

\usepackage{inconsolata}

\usepackage{graphicx}

\usepackage{xcolor}
\usepackage{xparse}
\usepackage{longtable}
\usepackage{array}
\usepackage{caption}
\usepackage{graphicx}
\usepackage{adjustbox}
\usepackage{booktabs}
\usepackage{balance}
\usepackage{tcolorbox}
\usepackage{adjustbox}
\usepackage{lipsum}

\definecolor{darkblue}{RGB}{0, 51, 160}
\definecolor{maroon}{RGB}{128, 0, 0}

\title{Abstracting Cross-Domain Action Sequences into Interpretable Workflows}

\author{Gaurav Verma \and Scott Counts \\
        Microsoft Corporation\\
        {\small\texttt{\{gauravverma,counts\}@microsoft.com}}
        }

\begin{document}
\maketitle
\begin{abstract}
Sequential or time‑stamped interaction logs provide objective records of digital application usage, yet their granularity and noise often obscure meaningful insights into people’s work. Such insights are essential for improving digital products in ways grounded in real‑world user interactions. Prior research has applied deep learning models to cluster user actions into high-level activities, but these approaches are highly sensitive to noise and struggle to generalize across applications. To address this limitation, we introduce WorkflowView, a framework that uses large language models (LLMs) to abstract low-level action sequences into high-level activities. We establish the effectiveness and generality of our approach across three distinct, challenging sequential tasks and diverse domains: (a) zero-shot task description reconstruction from browser logs (achieving high semantic similarity, $\mu_{sim} = 0.91$), (b) few-shot student dropout prediction using MOOC interaction logs (reaching weighted $F_1 = 0.90$ with only five few-shot examples), and (c) anonymized, privacy-preserving analysis of AI tool integration within document workflows in Microsoft Word. Our work demonstrates that LLM-based abstraction is a robust and efficient path forward for transforming low-level behavioral data into high-level, interpretable, and actionable insights. We also discuss practical considerations for deploying LLM‑based inferences within logging infrastructures, including computational efficiency and user privacy.
\end{abstract}

\section{Introduction}
\begin{figure}[!t]
    \centering
        \includegraphics[width=0.95\linewidth]{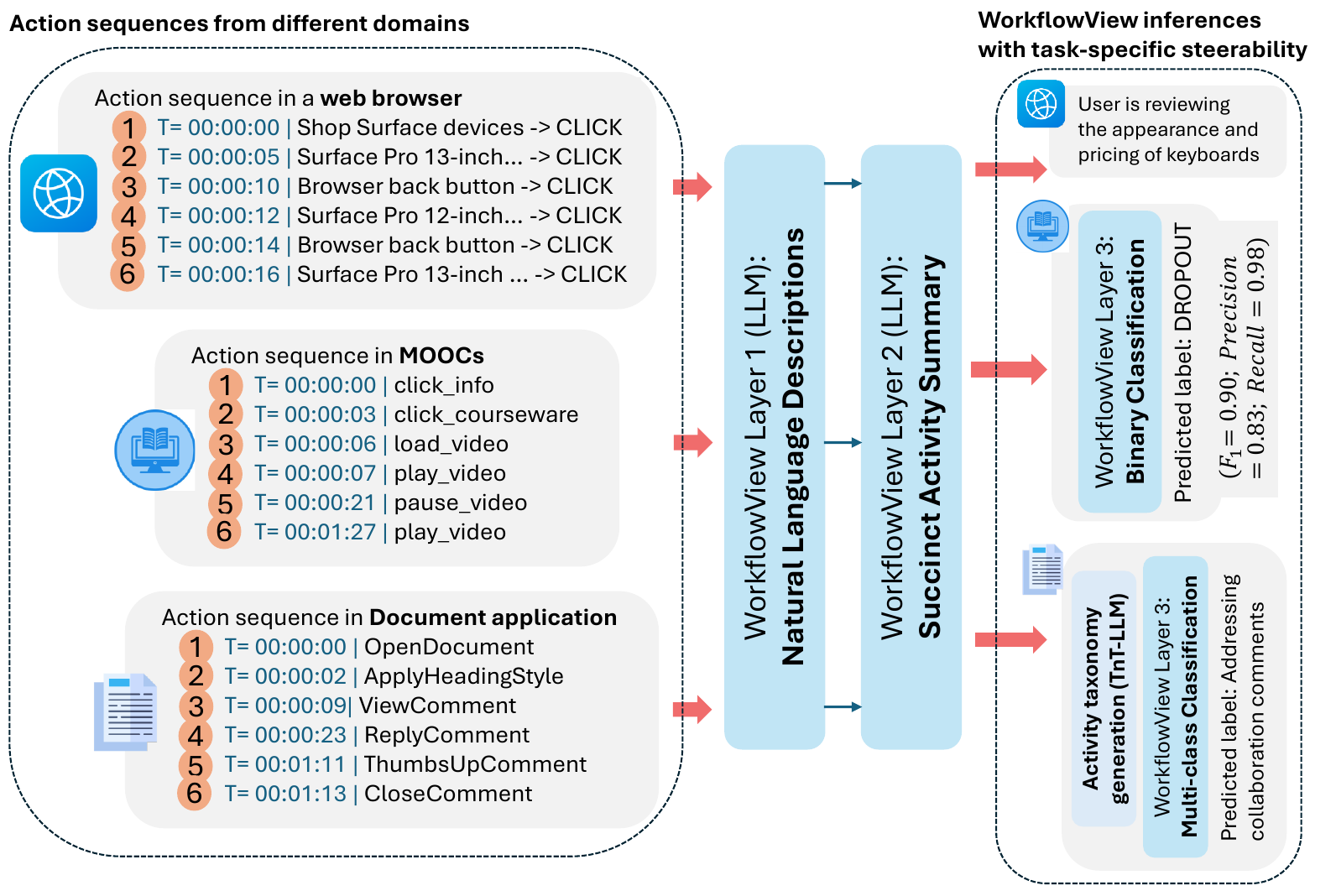}
        \caption{\small{We propose an LLM-based framework for hierarchical abstraction of user action sequences into interpretable high-level activities (WorkflowView). The left panel illustrates raw action sequences from three domains. WorkflowView enables downstream inferences with task-specific steerability, such as reconstructing user intent in browsers, predicting student dropout in MOOCs, and privacy-preserving categorization of document-centric workflows.}}
        \label{fig:overview-figure}
\end{figure}

Terabytes of user interface (UI) interaction logs are captured every hour as users interact with digital applications. These logs enable unobtrusive analysis of usage patterns, facilitate bug identification, and support iterative deployment of product improvements that better align with users’ needs. UI interaction logs provide an objective account of what actions users perform and when they perform them (e.g., \texttt{(DD/MM/YY HH:MM:SS, ClickedLayoutRibbon)}). However, such timestamped action sequences are often too granular and noisy to yield a clear view of the high-level task a user is performing within the application. A single high-level task (e.g., formatting the content of a document) may comprise hundreds of actions executed over a 10-15 minute interval, making the action sequence highly \textit{granular}. Moreover, these sequences may include actions that are not directly related to the user’s underlying intent, introducing \textit{noise}. For example, when users briefly click on unrelated features to intentionally or unintentionally explore the interface.

Earlier studies that model time-stamped interaction logs to understand user behavior have relied on statistical techniques such as frequent itemset mining and sequential pattern mining~\cite{mannila1997discovery, cuke2009new, agrawal1995mining, agrawal1993mining}. These approaches have been noted to struggle with incorporating domain context and with explicitly modeling noise in user behavior~\cite{dev2017identifying}. More recent work has explored adapting language modeling techniques to sequential log data; for example, using LSTMs to preemptively identify when users might need assistance within an application~\cite{nambhi2019stuck}, or training BERT- and LLM-based classifiers to detect anomalies in logs~\cite{guo2021logbert, zhou2024leveraging}. While these methods demonstrate the promise of interpreting interaction logs using language models, they typically operate in settings that require task-specific fine-tuning on thousands of annotated training samples.

Motivated by the strong generalization capabilities of large language models (LLMs) across tasks and domains, this work investigates whether state-of-the-art LLMs can interpret real-world timestamped action sequences that do not follow the usual syntax or semantics of natural language and infer the high-level activities that users perform as part of their workflows. LLMs are also known to integrate presented instances with broader encoded knowledge~\cite{bai2024kgquiz}, which may further enrich observability of system and user states. To this end, we propose WorkflowView, a hierarchical abstraction of granular action sequences using LLMs. In WorkflowView, the initial layer generates natural language descriptions of the observed actions, while subsequent layers infer high-level activities and, optionally, categorize them into a set of discovered or predefined categories. To demonstrate the generality of the proposed approach, we evaluate WorkflowView across three domains that differ in action set cardinality and in the degree to which user behaviors are mutually exclusive. Figure~\ref{fig:overview-figure} shows an overview of the method.

Our results show that WorkflowView provides a reliable abstraction over action sequences across diverse tasks and domains. Specifically, we find that the method \textit{(a)} generates task descriptions that closely align with ground-truth tasks performed in a browser (e.g., prediction: the user is trying to ``find a car while sorting by lowest price''; ground truth: the user wants to ``find the cheapest car''), \textit{(b)} predicts student dropout in MOOCs with a weighted $F_1$ score of 0.90 while using only five in-context examples (a performance comparable to several state-of-the-art predictive models trained on thousands of labeled instances), and \textit{(c)} contextualizes the use of AI tools in Microsoft Word (i.e., a document creation, collaboration, and consumption application) by interpreting action sequences, discovering task categories, and performing multi-class classification. We further show that such anonymous, privacy-preserving, and aggregated insights can inform user-centric product improvements.

Because WorkflowView relies on LLM-based inference over action sequences, we discuss practical considerations around deployment, including cost, latency, and user-privacy, as well as the limitations of our approach. We also outline a broader vision in which LLM capabilities are embedded deeper into the logging infrastructure. This vision is especially relevant in the context of human–AI collaboration, while maintaining strong guarantees around user privacy and security.

\section{Related work}
Below, we categorize and discuss related work into three themes: \textit{(a)} modeling interaction logs, \textit{(b)} discovering user intents from user utterances, and \textit{(c)} using LLMs to model non-language data.

\vspace{0.05in}
\noindent\textbf{Modeling interaction logs}:
Prior work on interpreting timestamped UI logs has largely framed the problem as pattern mining or sequence modeling. Techniques such as frequent itemset mining and sequential pattern mining have been widely used to extract common action patterns from large log corpora (e.g., identifying frequently occurring operation groups)~\cite{mannila1997discovery, cuke2009new, agrawal1995mining, agrawal1993mining}. While effective at identifying recurring structures, these statistical approaches are largely domain-agnostic: they treat UI actions as abstract tokens without semantic grounding~\cite{dev2017identifying} and are sensitive to noise and spurious correlations in action sequences~\cite{yang2002mining}. Subsequent work addressed some of these limitations through learning-based approaches, including RNN/LSTM- and Transformer-based models~\cite{hochreiter1997long, vaswani2017attention}, applied to domain- and task-specific applications~\cite{nambhi2019stuck, krishna2018lstm, zhu2021unilog}. However, these methods rely on task-specific training data and hand-crafted labels, making them costly to deploy across new domains, tasks, or evolving user behaviors. In contrast, WorkflowView relies on LLM-based inference via prompting, enabling flexible adaptation across tasks and domains without fine-tuning or annotating data, while explicitly abstracting away low-level noise through hierarchical reasoning.

\vspace{0.05in}
\noindent\textbf{Intent discovery over user utterances}:
A related line of work focuses on inferring user intent from textual interactions, such as search queries~\cite{wang2022recognizing} or conversational utterances in dialogue systems~\cite{schuurmans2019intent}. Modern dialogue systems and virtual assistants typically include an intent classification module that maps user input to predefined task labels (e.g., booking a flight or checking the weather), often trained using supervised learning over large annotated corpora~\cite{serban2015survey}. More recent work explores discovering new or evolving intents by clustering user queries that fall outside known categories~\cite{shah2025using, wan2024tnt}. A key distinction between this body of work and ours lies in the nature of the input: textual utterances are already semantic and human-interpretable, and often explicitly encode user goals (e.g., ``find the cheapest car'' or ``schedule a meeting''). In contrast, our work operates on telemetry data consisting of low-level UI events, where intent must be inferred indirectly from noisy, granular action sequences. This setting is both more challenging and more ubiquitous in modern applications, motivating the need for methods that can bridge raw interaction logs and high-level intent.

\vspace{0.05in}
\noindent\textbf{LLMs for non-language sequential data}:
Beyond text, recent work has examined the ability of LLMs to reason over non-language data. Existing approaches include learning projection layers to map image or numeric sensor data into representations suitable for LLM-based inference~\cite{verma2024cross, moon2024anymal}, adapting LLM embeddings for time-series classification~\cite{kaur2025lets}, and reprogramming time series into textual prototype representations that align more naturally with LLM pretraining~\cite{jin2023time}. Liu et al.~\citeyear{liu2024lstprompt} demonstrate that off-the-shelf LLMs such as GPT-4~\cite{achiam2023gpt} can outperform pre-trained zero-shot baselines (and, in many cases, supervised models) on forecasting numeric sequences across domains including epidemiology, finance, and weather. These results suggest that LLMs can, to some extent, interpret structured sequences with limited linguistic content by leveraging patterns learned during large-scale pretraining. Building on this insight, WorkflowView extends zero-shot and few-shot LLM prompting to the domain of user interaction logs.

\section{WorkflowView: Hierarchical abstraction of action sequences with LLMs}
WorkflowView is a simple yet effective framework that leverages large language models (LLMs) to reason over action sequences. The method demonstrates that LLMs can be prompted to address a range of sequence modeling tasks across domains in zero-shot or few-shot settings, highlighting ease of customization without fine-tuning. To encourage stage-wise abstraction from low-level actions to high-level activities, WorkflowView adopts a hierarchical design. We distinguish three levels of behavioral granularity: individual \textit{actions} (atomic UI events, such as a single click or keystroke); \textit{high-level activities} (coherent, interpretable units of behavior abstracted from a span of actions, such as ``reviewing comments''); and \textit{workflows} (goal-directed processes that these activities compose, such as collaborating on a document). Specifically, action sequences are first converted into detailed natural language descriptions (Layer 1), after which the high-level activity captured by these descriptions is inferred (Layer 2). If required by the task, additional layers can be introduced to further categorize the inferred high-level activity into known or discovered classes—for example, predicting student dropout in a MOOC or distinguishing between active document editing and text formatting. Figure~\ref{fig:overview-figure} provides an overview of the approach along with example outputs from the datasets used in this work.

The hierarchical LLM-based inference is motivated by two principles: \textit{modularity} and \textit{progressive denoising}. Modularity ensures that the outputs of lower layers (i.e., action sequence $\rightarrow$ natural language description $\rightarrow$ high-level task inference) can be reused across multiple objectives (such as frequent task discovery at the population level or categorization of individual sequences) by adapting only the higher layers. Progressive denoising is essential for modeling action sequences with LLMs, as it enables the transformation of raw timestamped actions into coherent textual representations that are better suited for higher-order reasoning. For instance, lower layers may capture temporal patterns in natural language, such as “the user responded to a collaborator’s comment after no significant activity for $N$ minutes.” In this case, low-salience actions are denoised at earlier layers, and depending on the value of $N$ (e.g., 2 \textit{vs.} 10 minutes), subsequent layers can characterize the level of deliberation involved in responding to the comment. See our discussion on the effectiveness of progressive denoising in Appendix \ref{appsec:additional-related-work}.

We provide the prompts used in WorkflowView in Appendix Tables~\ref{apptab:layer1-mind2web}, \ref{apptab:layer2-mind2web}, \ref{apptab:layer1-mooc}, and \ref{apptab:layer-mooc-summary} to support reproducibility and future work. In the following section, we evaluate WorkflowView on three tasks spanning three domains: inferring browser tasks, predicting student dropout in MOOCs, and contextualizing the use of an AI tool in Microsoft Word. Given the substantial variation in action spaces and behavioral patterns across these domains, our experiments are designed to evaluate WorkflowView's effectiveness and generalizability.

\section{Applications \& Evaluation of WorkflowView}
\subsection{Inferring tasks from browser interaction logs}
\label{sec:tasks-mind2web}
\noindent\textbf{Task and dataset}: We evaluate the ability of WorkflowView to infer the tasks people do on browsers using observed interaction logs alone. We use the Mind2Web dataset~\cite{deng2023mind2web}, which contains an ordered sequence of web actions taken in a browser to complete 2,022 general-purpose web tasks described in natural language. The tasks span 137 websites and 5 different domains: service (e.g., \textit{gov.uk}), shopping (e.g., \textit{instacart.com}), entertainment (e.g., \textit{espn.com}), travel (e.g., \textit{delta.com}), and information (e.g., \textit{finance.yahoo.com}).  The action space for this dataset is characterized by HTML UI elements (for instance, [button] `Go Back`, [textbox] `Enter your name') that the user interacts with on a webpage and the operation they perform (like CLICK, TYPE, or SCROLL). Our goal is to perform LLM abstractions over the action sequences, as exemplified in Table \ref{tab:mind2web-qual-example} (action sequences), to predict the task the users are doing across different websites (task). Methodologically, for this task, we operationalize WorkflowView using the prompts shown in the Appendix, where Layer 1 (shown in Table \ref{apptab:layer1-mind2web}) provides a detailed description of the action sequences in natural language and Layer 2 (shown in Table \ref{apptab:layer2-mind2web}) infers the overall task the user is doing and generates its succinct description. It is worth noting that this evaluation is a zero-shot setting. All our key results are based on experimentation with GPT-4o (more specifically \texttt{gpt-4o-2024-05-13}), a leading proprietary state-of-the-art large language model released by OpenAI~\cite{gpt4o}. However, we also demonstrate that WorkflowView works effectively with smaller \& open-weights models like Phi-4~\cite{abdin2024phi} and \texttt{gpt-oss-20b}~\cite{openai2025gptoss120bgptoss20bmodel} in App Table \ref{tab:mind2web-ranking-measures-phi}. 

\begin{table}[!t]
\centering
\begin{tabular}{l|c|c}
\textbf{Metric} & \textbf{Global} & \textbf{Website} \\
\toprule
\textbf{MRR} & 0.90 $(\pm 0.08)$ & 0.94 $(\pm 0.06)$ \\
\midrule
\textbf{Recall}@1 & 0.86 $(\pm 0.13)$ & 0.92 $(\pm 0.09)$ \\
Recall@3 & 0.94 $(\pm 0.07)$ & 0.98 $(\pm 0.04)$ \\
Recall@5 & 0.96 $(\pm 0.05)$ & 0.99 $(\pm 0.03)$ \\
Recall@10 & 0.98 $(\pm 0.01)$ & 0.99 $(\pm 0.01)$ \\
\end{tabular}
\caption{Embedding-based retrieval of ground-truth task descriptions using task descriptions generated using WorkflowView; the candidate set varies as `global' or `website-specific' across the two settings. $\mu (\pm \sigma)$.}
\label{tab:mind2web-ranking-measures}
\end{table}

\begin{table}[!b]
\centering
\begin{tabular} {p{0.14\textwidth} | p{0.30\textwidth}}
\toprule
\textbf{\small Action Sequence} & \tiny{[svg] → CLICK, [link] Your lists → CLICK, [link] Create a list → CLICK, [svg] → CLICK, [span] Walgreens → CLICK, [textbox] Add a title (Required) → TYPE: Walgreens, [img] → CLICK, [button] Next → CLICK, [link] Personal Care → CLICK, [svg] → CLICK, [img] → CLICK, [span] Add to list → CLICK, [checkbox] Walgreens New → CLICK, [button] Done → CLICK, [path] → CLICK, [path] → CLICK, [path] → CLICK, [svg] → CLICK, [img] → CLICK, [span] Add to list → CLICK, [checkbox] Walgreens New → CLICK, [button] Done → CLICK, [path] → CLICK, [link] View More → CLICK, [img] → CLICK, [span] Add to list → CLICK, [checkbox] Walgreens New → CLICK, [button] Done → CLICK, [button] Back → CLICK, [path] → CLICK, [link] Shower Essentials → CLICK, [img] → CLICK, [span] Add to list → CLICK, [checkbox] Walgreens New → CLICK, [button] Done → CLICK, [button] Back → CLICK, [link] Lists → CLICK}\\
\midrule
\textbf{\small Generated Task Description} & \small{Create a Walgreens shopping list and add personal care and shower essentials items.}\\
\midrule
\textbf{\small Ground-truth Task Description} & \small{Create a new list and add four items from the personal care category at Walgreens.} \\
\bottomrule
\end{tabular}
\caption{Qualitative example of task description generated using WorkflowView using the action sequence, and the corresponding ground-truth task description. }
\label{tab:mind2web-qual-example}
\end{table}

\vspace{0.05in}
\noindent\textbf{Evaluation settings}: As the first measure to compare the generated task descriptions from sequences of web interactions and their corresponding ground-truth descriptions, we compute the cosine similarity between the embeddings of the descriptions obtained from the \texttt{text-embedding-ada-002} model~\cite{embedding-model}.   Additionally, we compute retrieval metrics like Mean Reciprocal Rank (MRR) and Recall@$K$ ($K \in \{1, 3, 5, 10\}$) under two settings. In the first setting (i.e., `global'), we retrieve the most similar ground-truth task description across the entire dataset for each of the generated task descriptions; whereas, in the second setting (i.e., `website-specific'), we retrieve the most similar ground-truth description across a website for each of the generated descriptions belonging to the same website. 

\vspace{0.05in}
\noindent\textbf{Results}: The average similarity (and standard deviation) between the generated and ground-truth tasks is $0.911$ $(\pm$ $0.042)$; $N = 2,022$ tasks. The notably high absolute similarity scores indicate the accurate inferences made using WorkflowView. Table \ref{tab:mind2web-ranking-measures} shows the average MRR and Recall@$K$ values (along with standard deviations). The near-perfect MRR and Recall@K values also indicate that the true ground-truth descriptions are ranked at the top for a large majority of the corresponding generated task descriptions. In Table \ref{tab:mind2web-qual-example}, we qualitatively illustrate the close alignment between a generated task description and the corresponding ground-truth description; an expanded set of qualitative examples is present in Appendix Table \ref{tab:mind2web-qual-examples}. The key strength of our work lies in demonstrating zero-shot, cross-domain generality of WorkflowView; nonetheless, we compare against domain-specific fine-tuned seq2seq~\cite{sutskever2014sequence} baselines for this task in Appendix \ref{appsec:baselines-mind2web}.  

\subsection{Predicting student dropouts based on MOOC interaction logs}

\begin{figure*}
    \centering
    \includegraphics[width=1.0\linewidth]{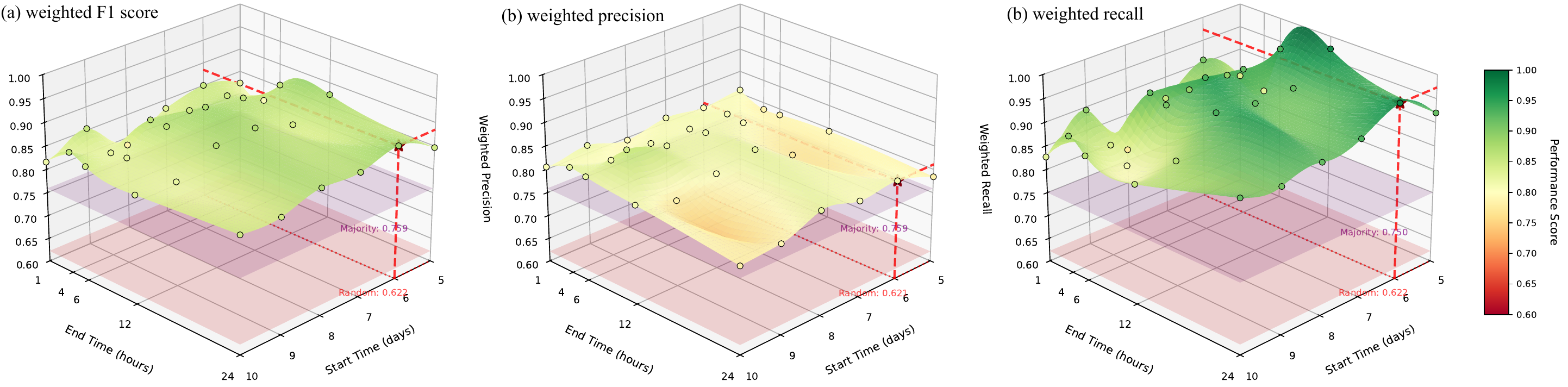}
    \caption{\small{Dropout prediction performance using WorkflowView. The plots show the weighted F$_1$ score, precision, and recall in response to the variations in `start time' (i.e., time when the action sequences are modeled) and  `end time' (i.e., time before the last action until which the action sequences are modeled). For comparison, we include the scores corresponding to baselines where only the `majority' class would be predicted (i.e., all dropout) and predictions based on biased `random' guesses as per prior class probabilities. The best $F_1$ score ($F_1$ = 0.89; Precision = 0.81; Recall = 0.97) correspond to a start time of 6 days and an end time of 24 hours. The number of few-shot examples provided to WorkflowView were 3 for this analysis; Figure \ref{fig:mooc-few-shot-sensitivity} below shows the sensitivity to the number of few-shot examples.}}
    \label{fig:mooc-time-analysis}
\end{figure*}

\begin{figure}
    \centering
    \includegraphics[width=1.0\linewidth]{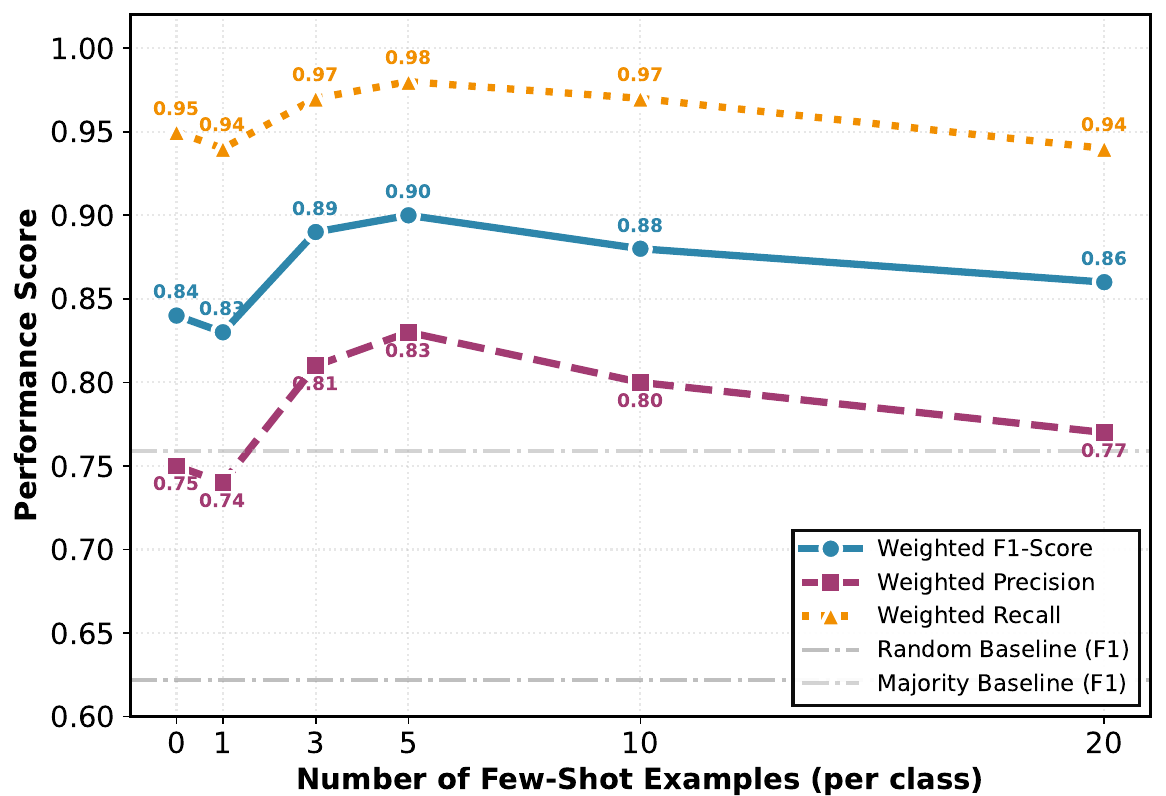}
    \caption{\small{Variation in predictive performance of WorkflowView (weighted F$_1$, precision, and recall) on the MOOC dropout prediction task, in response to the number of few-shot examples considered ($N \in \{0, 1, 3, 5, 10, 20\}$).}}
    \label{fig:mooc-few-shot-sensitivity}
\end{figure}

\noindent\textbf{Task and dataset}: To assess the method's generalizability across diverse tasks and domains, we experiment with interaction logs of a MOOC software to predict student dropouts.  The test set of the dataset curated by \citeauthor{feng2019dropout} (\citeyear{feng2019dropout}) comprises interaction logs from a total of 44,008 unique students enrolled in 247 unique courses, resulting in 67,699 unique (student, course) pairs. Of the 67,699 unique enrollments,  51,316 (75.8\%) resulted in a dropout. 22 unique actions were logged from all the student interactions, representing the action space. The goal of this task is to process the time-stamped sequence of actions at least N hours \textit{before the last action} using WorkflowView to determine if the enrollment is going to result in a dropout, such that $N \in \{1, 6, 12, 18, 24\}$ hours. The design for this predictive task takes into account a potential intervention to take place when the student performs their currently last action that may discourage them from dropping out. 

\vspace{0.05in}
\noindent\textbf{Adapting WorkflowView}: For this task, to facilitate a binary classification, we adapt WorkflowView to have a third categorization layer on top of the first two layers (natural language description and succinct summary). Effectively, if the final binary classification labels are accurate, it indicates that WorkflowView can interpret and extract meaningful task-specific signals from the raw action sequences. This particular task and the low-barrier adaptation of WorkflowView to address it emphasizes the modularity of the underlying hierarchical abstractions. We also explore WorkflowView's compatibility to few-shot settings, by modifying the prompts at each layer to include illustrative mappings. Specifically, for the natural language description layer (Layer 1) this was done by providing the mapping between action sequences and the final category; for the succinct summary layer (Layer 2) this was done by additionally including the natural language descriptions from the previous layer for the same examples, and similarly, for Layer 3, we additionally included the succinct summaries for the examples. The prompts used to adapt WorkflowView for this task are shown in the Appendix Tables \ref{apptab:layer1-mooc}  and \ref{apptab:layer-mooc-summary}. We explored few-shot settings where the number of examples per category varied in $\{1, 3, 5, 10\}$, where that many examples were randomly sampled from the training set per category (i.e., the total number of examples were twice as many).

\begin{figure*}[!h]
    \centering
    \begin{minipage}[!t]{0.49\textwidth}
        \centering
        \includegraphics[width=\linewidth]{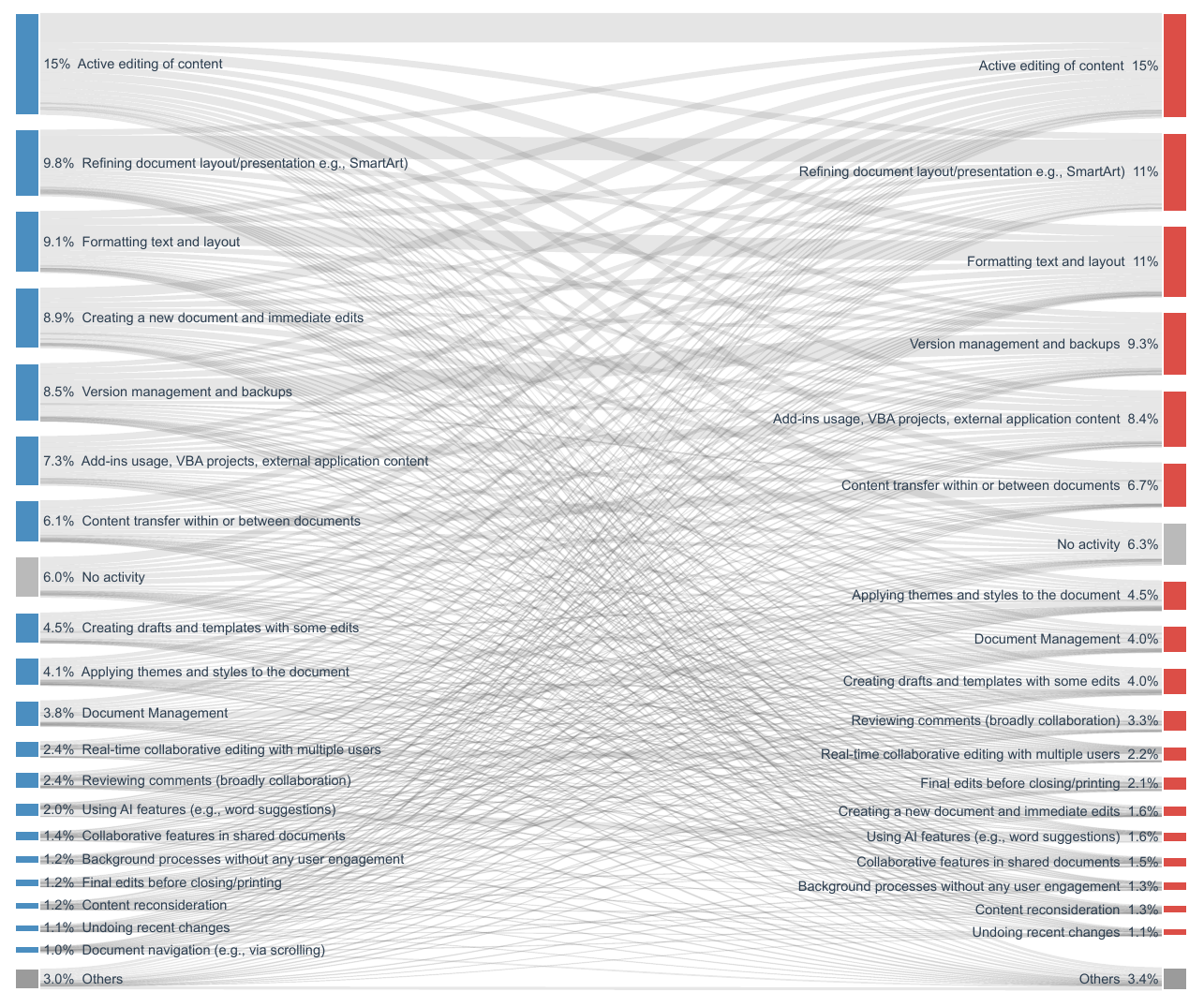}
        \caption{\small Activities users do in the context of document editing, at most 30 minutes before (left, in blue) and after (right, in red) prompting the integrated AI tool and accepting its output. The action sequences are processed using WorkflowView to discover the high-level categories; corresponding definitions are provided in the adjoining table.}
        \label{fig:workflow-sankey}
    \end{minipage}%
    \hfill
    \begin{minipage}[!t]{0.49\textwidth}
        \centering
        \tiny{
        \begin{tabular}{p{0.95\textwidth}}
        \textbullet\hspace{0.5mm}\textbf{Active editing of content}: Modifying the content of a document, such as copying, pasting, deleting, and reorganizing text. \\
        \textbullet\hspace{0.5mm}\textbf{Refining document layout/presentation}: Refining and enhancing document layout and presentation, including inserting graphics. \\
        \textbullet\hspace{0.5mm}\textbf{Formatting text and layout}: Changing the appearance of text and its layout, including font adjustments, applying styles, and using tools like Format Painter. \\
        \textbullet\hspace{0.5mm}\textbf{Add-ins usage, VBA projects, external application content}: Interactions with external applications or add-ins within the context of document editing. \\
        \textbullet\hspace{0.5mm}\textbf{Version management and backups}: Create different versions or backups of a document. \\
        \textbullet\hspace{0.5mm}\textbf{Creating a new document and immediate edits}: Starting a new document and immediately engaging in editing or content insertion. \\
        \textbullet\hspace{0.5mm}\textbf{Content transfer within or between documents}: Copying and pasting content within or between documents. \\
        \textbullet\hspace{0.5mm}\textbf{Applying themes and styles to the document}: Organizing and structuring document content, including applying themes and styles. \\
        \textbullet\hspace{0.5mm}\textbf{Creating drafts and templates with some edits}: Creating drafts or templates, making edits, and possibly using draft generation features. \\
        \textbullet\hspace{0.5mm}\textbf{Reviewing comments (collaboration)}: Engaging with comments, indicating a review or collaboration phase. \\
        \textbullet\hspace{0.5mm}\textbf{Real-time collaborative editing with multiple users}: Engaging in real-time collaboration and editing with others. \\
        \textbullet\hspace{0.5mm}\textbf{Final edits before closing/printing}: Preparing the document for presentation or distribution, including final edits, formatting, and printing. \\
        \textbullet\hspace{0.5mm}\textbf{Collaborative features in shared documents}: Shared document activities, including co-authoring, managing comments, and using collaborative features. \\
        \textbullet\hspace{0.5mm}\textbf{Using AI features (e.g., word suggestions)}: Using AI features to edit the document content. \\
        \textbullet\hspace{0.5mm}\textbf{Content reconsideration}: Experimenting with content by adding and then removing it, indicating reconsideration of content placement or inclusion. \\
        \textbullet\hspace{0.5mm}\textbf{Document navigation}: Navigating through the document, including moving the cursor or scrolling. \\
        \end{tabular}
        }
        \label{tab:doc-category-definitions}
        \captionof{table}{\small Discovered document editing categories and their corresponding descriptions.}
    \end{minipage}
    \label{fig:sankey-visual-and-definitions}
\end{figure*}

\vspace{0.05in}
\noindent\textbf{Evaluation setup}: The evaluation is designed to measure how effectively and reliably can WorkflowView perform the binary classification task of predicting student dropouts from MOOC action sequences. Our evaluations are centered around two axes that could precipitate predictive variability: \textit{(a)} time horizon for the action sequences under consideration, and \textit{(b)} the number of few-shot examples provided to the model. For the former, we vary the start time as well as the end time and observe the weighted $F_1$ score, precision, and recall for each combination. For the latter, we explore both zero-shot and few-shot settings, while varying the number of examples considered in the few-shot setting. In the few-shot setting, we randomly sample $K$ examples per category from the train set of the data; we acknowledge that sampling strategies that result in better performance and greater robustness are possible~\cite{wang2020generalizing,nookala2023adversarial}. To limit the amount of experimentation we first evaluate the performance of WorkflowView on a 2-dimensional hyper-parameter grid of start and end times while fixing the number of few-shot examples to 3, and then evaluate the sensitivity to the number of few-shot examples.

\vspace{0.05in}
\noindent\textbf{Results}: Figure \ref{fig:mooc-time-analysis} shows the predictive performance in response to variations in start and end time hyper-parameters. For reference, we also include comparisons with two random baselines: `majority', where only the majority class is predicted (i.e., all sequences are categorized as dropout) and `random', where the categorizations are based on class probabilities based on the training data distributions. The first key observation is that regardless of the start and end times hyperparameters, WorkflowView categorizations are consistently and notably better than either of the baselines. In fact, the best performance across all the combinations comes out at an $F_1$ score of $0.89$ (Precision $= 0.81$ and Recall $= 0.97$) corresponding to a start time of 6 days and an end time of 24 hours before the last activity. It is worth noting that this performance is on par with several learning-based methods that utilize over hundreds of thousands of training examples. \footnote{\citeauthor{fu2021clsa} (\citeyear{fu2021clsa}) train a long short-term memory network to build a predictive model that achieves an $F_1$ score of 0.869 on the binary classification task of MOOC dropout prediction. Similarly, \citeauthor{basnet2022dropout} (\citeyear{basnet2022dropout}) propose training-based approaches that rely on large-scale annotated data and result in an $F_1$ score of 0.84; \citeauthor{feng2019dropout} (\citeyear{feng2019dropout}) report an $F_1$ score of 0.91. See App. \ref{appsec:baselines-mooc} for baseline comparisons.} Next, we fix the start time of the action sequence used for predictive modeling to 6 days and the end time to 24 hours, and vary the number of few-shot examples supplied to the model. Figure \ref{fig:mooc-few-shot-sensitivity} shows that using 3 or 5 few-shot examples per category improves the performance considerably over the zero-shot setting ($F_1$ improves from $0.84$ to $0.89$ and $0.90$, respectively). The minor drop in performance with only a single few-shot example can be explained by the two operating modes of in-context learning~\cite{lin2024dual, min2022rethinking}: with insufficient demonstrations, models tend to rely on retrieving familiar tasks from pretraining (`task retrieval') rather than adapting to the presented task (`task learning'). However, on further increasing the number of in-context learning examples from $5 \rightarrow 10 \rightarrow 20$, the performance drops possibly because of increasing context length. More broadly, applying WorkflowView to predict student dropouts based on MOOC interaction logs not only indicates the effectiveness of the method regardless of the hyperparameters related to sequence length and duration but also provides insights into how the method interacts with the broader literature around in-context learning.

\subsection{Analyzing before/after activities around key actions: A case-study of how AI tools help in document workflows}  As AI assistance tools are transforming how users engage with documents across different digital applications, we conduct a case-study on analyzing the action sequences before and after users accept the response provided by an AI tool (Copilot in Word; as captured by a specific action in the application telemetry). The AI tool considered in this study is embedded as a product feature within Microsoft Word, a document editing application with hundreds of millions of active users. The application allows the users to prompt the AI tool at any stage of their workflow while working with a document and accept or discard its presented output. The case-study demonstrates how WorkflowView could provide interpretable and actionable user-centric insights that could improve interactions and product design. 

We used WorkflowView to analyze the anonymous and privacy-preserving telemetry of users of Microsoft Word. We sampled $50,000$ users who had interacted with the AI tool at least once in the month of June 2025 and were located in the United States with their application language set to \texttt{us-en}. Users consented to log collection as part of the user agreement. Note that the interaction logs are devoid of textual data and writer data, and the telemetry that captures users' interactions with the application UI is highly granular and include approximately 2000 unique actions. We only present aggregated, percentage-based insights over the random sample of the users. 

Two key modifications exist beyond applying Layers 1 (i.e., natural language descriptions) and 2 (succinct activity summary) of WorkflowView for this task. Since there is no list of prior activities that the action sequences need to be mapped to, these activities have to be `discovered' from the data. Once these activities have been discovered, there is a need for a categorization layer that maps the succinct summaries to one of the discovered categories.  For the category discovery step, we use an existing method (TnT-LLM~\cite{wan2024tnt}) that does end-to-end label generation based on the raw succinct activity summaries (i.e., Layer 2's output). The identified labels are then used in Layer 3 of WorkflowView for multi-class classification, akin to the binary classification task for MOOC dropout prediction. We included the description of the high-level categories that were obtained using TnT-LLM in the previous step to inform the multi-class classification. This case-study also illustrates another example of easy adaptation of WorkflowView to applications that may involve identifying categories of activities or model evolving user behavior, where existing categories may become outdated over time.

\vspace{0.01in}
\noindent\textbf{Dataset}:  We processed the sequences before and after (at most 30 minutes in duration) each of the occurrences of the action that indicates keeping AI tool's output using WorkflowView. Finally, for discovering the category of activities using TnT-LLM, we used $20\%$ of the occurrences and then inferred the categories (using Layer 3) on the entire action sequence set. 

\vspace{0.01in}
\noindent\textbf{Analysis and Insights}: Figure \ref{fig:workflow-sankey} shows that active content editing (described as activities involving modifying content of a document, such as copying, pasting, deleting, and reorganizing text in Table 3) is the most frequent activity both before (15\%) as well as after (15\%) AI assistance, indicating its prominence in document-related workflows where AI tools are used. Active editing of content tends to continue as such after accepting AI tool's outputs or, in certain cases, the user tends to transition to other activities like formatting text and its layout or transferring content within or across documents. It is worth noting that the share of activities pertaining to formatting or refining layouts is greater \textit{after} the AI tool's responses are accepted when compared to their share before, which may indicate that users try to incorporate the AI tool's output in a manner that is consistent with the original content's formatting. The insights obtained with WorkflowView enable interpreting user engagement patterns from noisy and granular action sequences. Additionally, these insights also offer actionable guidance for product improvements such as introducing more context-aware formatting suggestions or adaptive layout tools that align with post-AI interaction behaviors.

It is also worth noting that in an evolving landscape where AI tools are changing how users interact with applications, it is a strength that the activities (and the corresponding descriptions)  are \textit{synthesized} with the activity summaries inferred by WorkflowView from the action sequence data, rather than being predefined labels. This ensures that the taxonomy reflects actual user behavior as it evolves over time. We discuss practical considerations around efficiency in the context of real-world deployment in the following section. 

\section{Discussion: Deployment and Extensions}
WorkflowView is an LLM-powered approach to do hierarchical abstractions over action sequences to understand users' behavior within digital applications. We demonstrate that the method can be easily adapted to work with a diverse set of tasks involving action sequences (task description generation, binary classification, category discovery and multi-class classification) across different domains (browser, MOOC, and document editing application). Quantitative and qualitative analyses demonstrate that WorkflowView is on par with training-based models for these tasks in zero-shot or few-shot settings. The results indicate the promise of embedding LLM-powered inferences in the lowest level of data infrastructure to drive user-centric product improvements. Below, we discuss some of the future extensions and applications of WorkflowView. 

\vspace{0.01in}
\noindent\textbf{Deployment cost and latency}:
Relying on LLM-based inference requires careful consideration of deployment cost and latency, particularly for applications with large user bases where interaction logs may span terabytes of data. Two trends are worth noting. First, the cost and latency of LLM inference have decreased rapidly in recent years~\cite{llminferencelatency,epoch2025llminferencepricetrends}. In parallel, there has been growing support for deploying smaller language models, which our evaluations suggest can perform on par with larger models for activity inference tasks. For example, Phi-4 (14B parameters;~\cite{abdin2024phi}) achieves performance comparable to GPT-4o on the web browser inference task while requiring substantially fewer hardware resources (Appendix Table~\ref{tab:mind2web-ranking-measures-phi}). Together, these trends make it increasingly feasible to apply LLM-based methods such as WorkflowView to large-scale interaction logs. In the near term, WorkflowView can be deployed in offline settings to help developers understand how users interact with their applications and to identify opportunities for product improvement. Periodic offline analyses can also surface shifts in user behavior over time, which is particularly relevant in dynamic human–AI collaborative workflows. Such deployments can further control cost by operating on representative samples of users, while avoiding the latency constraints associated with real-time inference.

\begin{figure}[!t]
\centering
\includegraphics[width=1.0\linewidth]{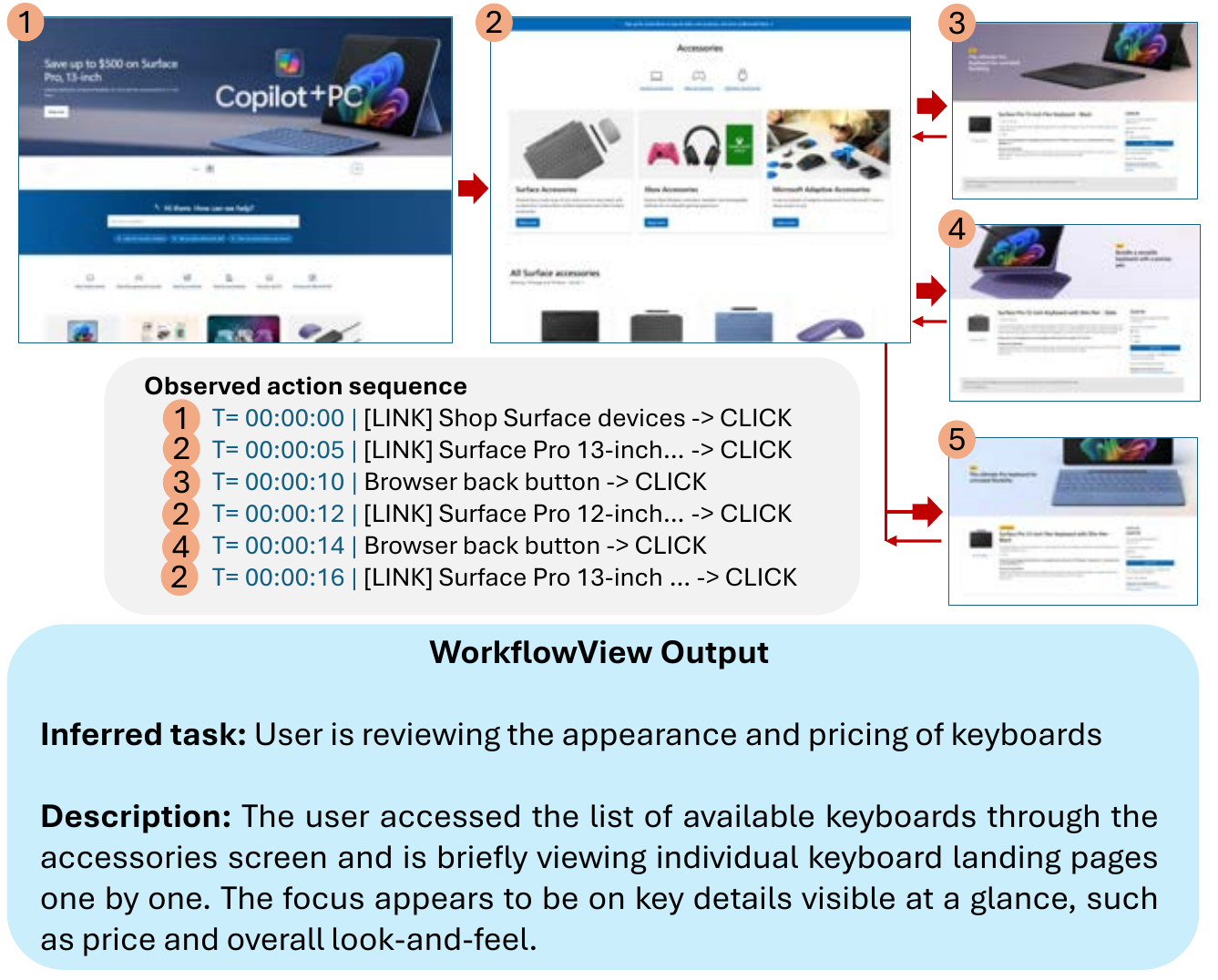}
\caption{\small Illustrative multimodal extension of WorkflowView for task inference using browser snapshots and UI click logs. In this example, task inference is performed unobtrusively as the user interacts with the application. The visual modality complements the textual action sequence by grounding descriptions in on-screen context (e.g., visible content and time spent), leading to more accurate and informative inferences than action sequences alone.}
\label{fig:multimodal-workflowview}
\end{figure}

\vspace{0.01in}
\noindent\textbf{Multimodal extensions of WorkflowView}: In this work, WorkflowView operates on action sequences captured in text. However, given the multimodal capabilities of modern LLMs, the framework can be naturally extended to incorporate UI screenshots captured at key transition points during user interaction. Visual context can provide complementary signals about user behavior, reduce reliance on application-specific instrumentation, and ground textual descriptions in the actual interface state. Figure~\ref{fig:multimodal-workflowview} qualitatively illustrates such a multimodal setting, where both high-level task inference and behavioral descriptions are captured accurately.

Multimodal extensions of WorkflowView open the door to real-time, proactive AI assistance that supports users in achieving their high-level goals. For example, a user browsing an e-commerce website could be offered structured recommendations—such as items to compare, stratified by viewed and unviewed—based on their inferred goal (e.g., evaluating alternative keyboards) and current progress toward that goal. More broadly, accurate real-time modeling of user behavior, encompassing both intent and task progress, is likely to be foundational for human–AI collaboration, enabling seamless hand-offs between users and AI systems.

\section{Conclusion}
Building on the generalization capabilities of LLMs, including their demonstrated effectiveness on non-language data, we introduce WorkflowView, a framework for hierarchical abstraction over action sequences to infer users' high-level activities within digital applications. We show that WorkflowView can be applied reliably across three diverse domains and across multiple tasks. Through a case study on real-world telemetry from Microsoft Word, we illustrate how WorkflowView enables anonymous, privacy-preserving, and aggregated analysis of user behavior that can inform user-centric product improvements. Finally, we outline multimodal extensions of WorkflowView that can support more effective human–AI collaboration and discuss key deployment considerations.


\vspace{-1mm}
\section{Limitations}
\vspace{-1mm}
\noindent\textbf{Privacy and security considerations}:
It is critical to consider the privacy and security implications associated with the design, deployment, and future extensions of WorkflowView. First, UI action sequences should only be collected with informed user consent, and inferences should be limited to behavioral understanding that does not reveal PII or sensitive content—for example, inferring that a user is ``actively applying formatting changes to text'' rather than ``actively formatting text in a legal contract.'' For real-time assistance, particularly in multimodal settings where private data could appear in UI screenshots, operationalization could involve strictly performing \textit{on-device} inferences, while logging only privacy- and security-compliant textual abstractions server-side for offline analysis that informs product improvements. Concrete privacy budgets or differential privacy (DP)-style guarantees could be explored in future work. Transparent and informed user consent is essential to ensure trust in AI-powered technologies.

\vspace{0.05in}
\noindent\textbf{Limitations}: First, the action names that make up action sequences must convey meaningful information about user interactions (e.g., `ClickLayoutRibbon' rather than `Action1'). This limitation also highlights the importance of developing an informative logging infrastructure to fully leverage LLM capabilities. Second, future work could explore more token-efficient prompting mechanisms to represent raw action sequences, as our current approach uses a direct textual representation of time-stamped actions. Simple strategies, such as chunking actions based on temporal proximity, could substantially reduce token counts. Finally, while this work evaluates off-the-shelf LLMs under zero-shot and few-shot settings with a focus on cross-task and cross-domain generalizability, future research could investigate large-scale pre-training on action sequences from diverse domains to further improve generalization across both in-sample and out-of-sample tasks.

\vspace{0.05in}
\noindent\textbf{Data}:
Two of the datasets were curated by prior work are publicly available~\cite{deng2023mind2web, feng2019dropout}; we comply with their terms of use. Users of Microsoft Word consented to interaction log collection as part of the user agreement.

\vspace{-1mm}
\section*{Disclaimer}
\vspace{-1mm}
Some of the information in this document relates to pre-released content which may be subsequently modified. Microsoft makes no warranties, express or implied, with respect to the information provided here. This document is provided ``as-is''. Information and views expressed in this document, including URL and other Internet Web site references, may change without notice. Some examples depicted herein are provided for illustration only and are fictitious. No real association or connection is intended or should be inferred. This document does not provide you with any legal rights to any intellectual property in any Microsoft product.

\balance
\bibliography{custom}

\appendix

\section{Appendix}
\label{sec:appendix}

\subsection{Comparison against fine-tuned baselines}
\label{appsec:fine-tuned-baselines}
While the core value proposition of WorkflowView is its zero-shot, cross-domain applicability, for completeness, we benchmark the performance against domain-specific fine-tuned baselines. For the browser task inference dataset, we compare against LSTM and BERT-based variants of sequence-to-sequence models~\cite{sutskever2014sequence}. For the MOOC dropout prediction, we compare against several approaches proposed in prior work that use different feature sets to perform the binary classification task, while keeping the evaluation settings consistent.

\subsubsection{Baselines for browser task inference}
\label{appsec:baselines-mind2web}
Since generating the task description using browser action sequences is a sequence-to-sequence task (akin to neural machine translation), we use the existing seq2seq models and their implementations. Using the train set of the Mind2Web dataset, the models learn the transformation of action sequences to the task description (word by word), such that actions are demarcated using the \texttt{[ACTION]} token. We use word2vec~\cite{mikolov2013efficient} embeddings, while randomly initializing the out-of-vocabulary words and keeping the embeddings trainable (as some of the interaction log vocabulary is not aligned with conventional language vocabulary). To avoid extensive hyperparameter tuning, we follow the training recipe and best practices described by ~\citeauthor{britz2017massive} (\citeyear{britz2017massive}) for LSTM-seq2seq and ~\citeauthor{rothe2020leveraging} (\citeyear{rothe2020leveraging}) for BERT-seq2seq closely; code available at \url{https://github.com/google/seq2seq} and \url{https://github.com/google-research/google-research/tree/master/bertseq2seq}, respectively. Our evaluations (shown in Table \ref{tab:mind2web-appendix}) are centered around the same metrics for the `Global' setting that is described in Section \ref{sec:tasks-mind2web}.

\begin{table}[!t]
\centering
\begin{tabular}{l cc}
\toprule
\textbf{Models} & \multicolumn{2}{c}{\textbf{Metrics}} \\
\cmidrule(lr){2-3}
& \textbf{MRR} & \textbf{Recall@1} \\
\midrule
LSTM seq2seq & 0.54 $(\pm 0.19)$ & 0.49 $(\pm 0.21)$ \\
BERT seq2seq & 0.68 $(\pm 0.16)$ & 0.65 $(\pm 0.18)$ \\
WorkflowView & 0.90 $(\pm 0.08)$ & 0.86 $(\pm 0.13)$ \\
\bottomrule
\end{tabular}
\caption{Comparing our training-free, zero-shot approach (WorkflowView) with domain-specific fine-tuned baselines for task inference using browser action sequences. }
\label{tab:mind2web-appendix}
\end{table}

\subsubsection{Baselines for MOOC dropout prediction}
\label{appsec:baselines-mooc}
We compare against the Context-aware Feature Interaction Network (CFIN) introduced by \citet{feng2019dropout}. CFIN models each enrollment using two feature groups: (i) learning-activity features $X(u,c)$ extracted from historical logs (primarily statistics over student actions), and (ii) context features $Z(u,c)$ capturing user and course attributes (e.g., demographics and course category). CFIN combines context-aware smoothing and feature-interaction modeling, and uses a \emph{3-layer} deep neural network (DNN) classifier as its prediction head. As their strongest variant, \citeauthor{feng2019dropout} propose an ensemble strategy (``CFIN-en'') analogous to stacking: they take the representation from the penultimate DNN layer and train an XGBoost classifier jointly on this representation and the original features $(X, Z)$. We use the authors' public implementation {\url{https://github.com/wzfhaha/dropout_prediction}} and re-train CFIN and the XGBoost-stacked ensemble using features constructed under our 6-day input/24-hour evaluation window, while otherwise following the paper/code defaults. We additionally compare against recent deep learning baselines that operate on week-level temporal feature vectors. Following \citet{yang2024deep} (CNN-LSTM) and its bi-attention variant (CNN-LSTM Bi-Att). We keep the architecture and hyperparameters consistent with the original work (e.g., dropout and early stopping), and reconstruct the  inputs to match our 6-day input/24-hour evaluation window. Table \ref{appsec:baselines-mooc} shows the comparison; WorkflowView achieves its best performance with 5 in-context examples per class (10 total), reaching a weighted $\mathcal{F}_1$ of 0.90. It is noteworthy that the performance is competitive even at 0-shot and remains stable across a range of fewshot budgets (see Fig. \ref{fig:mooc-time-analysis} \& \ref{fig:mooc-few-shot-sensitivity}).

\begin{table}[!t]
\centering
\begin{adjustbox}{max width=0.5\textwidth}
\begin{tabular}{l c}
\toprule
\textbf{Models} & \textbf{Weighted ${F}_1$} \\
\midrule
DNN (3-layer MLP) \cite{feng2019dropout} & 0.83 \\
CFIN-en \cite{feng2019dropout} & 0.90 \\
CNN-LSTM \cite{yang2024deep} & 0.86 \\
CNN-LSTM Bi-Att \cite{yang2024deep} & 0.87 \\
WorkflowView (0-shot) & 0.84 \\
WorkflowView (5-shot) & 0.90 \\
\bottomrule
\end{tabular}
\end{adjustbox}
\caption{Comparing our few-shot adapted approach to domain-specific supervised baselines for MOOC dropout prediction. Reported values are $\mu (\pm \sigma)$.}
\label{tab:mooc-appendix}
\end{table}

\subsubsection{Experiments with smaller LLMs}

\begin{table}[!h]
\centering
\begin{tabular}{lcc}
\toprule
\textbf{Metric} & \textbf{Phi-4 (14b)} & \textbf{GPT-OSS-20b} \\
\midrule
\textbf{MRR} & 0.89 $(\pm 0.09)$ & 0.90 $(\pm 0.08)$ \\
\textbf{Recall}@1 & 0.85 $(\pm 0.12)$ & 0.86 $(\pm 0.13)$ \\
Recall@3 & 0.93 $(\pm 0.05)$ & 0.94 $(\pm 0.07)$ \\
Recall@5 & 0.95 $(\pm 0.06)$ & 0.96 $(\pm 0.05)$ \\
Recall@10 & 0.97 $(\pm 0.02)$ & 0.98 $(\pm 0.01)$ \\
\bottomrule
\end{tabular}
\caption{Embedding-based retrieval of ground-truth task descriptions using task descriptions generated using WorkflowView. Phi-4 and \texttt{gpt-oss-20b} were used to generate the task descriptions.}
\label{tab:mind2web-ranking-measures-phi}
\end{table}

We evaluate the performance of smaller LLMs on browser task inference (`Global' setting; see Section \ref{sec:tasks-mind2web}) and consider two models --- Phi-4 (14B parameters)~\cite{abdin2024phi} and \texttt{gpt-oss-20b}~\cite{openai2025gptoss120bgptoss20bmodel}. This is largely to assess whether the LLMs that are among the fastest (as measured by tokens per second) and cheapest (cost per token)~\cite{open_llm_leaderboard_2025}, can also interpret user action sequences as well as more expensive counterparts like GPT-4o. We find that the task descriptions generated using Phi-4 and \texttt{gpt-oss-20b} demonstrated a mean similarity of $0.902 \pm 0.036$ and $0.909 \pm 0.039$, respectively. Table \ref{tab:mind2web-ranking-measures-phi} shows that the retrieval-based metrics are also on par with those obtained using the GPT-4o model in Table \ref{tab:mind2web-ranking-measures}. 

\subsection{Additional related work}
\label{appsec:additional-related-work}
\noindent\textbf{Hierarchical action modeling}: Beyond the pattern mining techniques discussed in Section 2, our work shares conceptual roots with hierarchical representation learning for sequential data. Traditional approaches in robotics and plan recognition have long used Hierarchical Hidden Markov Models (HHMMs) ~\cite{fine1998hierarchical} to decompose complex behaviors. In the human-computer interaction community, GOMS modeling ~\cite{card2018psychology} provided early foundations for decomposing user goals into operators, which WorkflowView's context involves replacing manual task analysis with LLM-based inference.

\vspace{0.02in}
\noindent\textbf{LLMs for system logs}: While we discuss LLMs for user action sequences, there is a growing body of work specifically targeting system logs for anomaly detection and root cause analysis. Methods like LogPPT ~\cite{le2023log} and Log-LLM ~\cite{ji2025adapting} demonstrate the utility of prompt-based learning for structured log parsing. WorkflowView differs by focusing on semantic user intent across diverse UI domains rather than system-level health monitoring.

\vspace{0.02in}
\noindent\textbf{In-context learning for sequence tasks and hierarchical prompting}: The sensitivity analysis in Section 4.2  regarding few-shot examples aligns with recent findings on the ``lost in the middle'' phenomenon and context window saturation ~\cite{liu2024lost}. More recently, the ``least-to-most'' prompting paradigm ~\cite{zhouleast} demonstrated that LLMs are significantly more effective when complex problems are decomposed into a series of simpler sub-problems, with the solution to each step facilitating the next. WorkflowView translates this principle to the domain of telemetry by treating raw event logs as the input ``complex problem'' and using a layered architecture to progressively abstract user intent. The current study does not include a formal ablation on the hierarchical structure itself (e.g., comparing single-pass prompting against the multi-layered approach). We maintain that this hierarchical design is central to the ``progressive denoising'' required for noisy and granular telemetry. As least-to-most prompting has already been shown to significantly outperform single-pass reasoning in tasks requiring complex decomposition and easy-to-hard generalization, we chose to implement this proven hierarchical logic rather than re-evaluating its effectiveness ablations.

\subsection{Quality \& stability of inferred activities}
It is important to address the stability of our unsupervised categorization, particularly within the Microsoft Word case study involving enterprise document software. Because this domain lacks a labeled ground-truth for inferred high-level categories, we conducted a sensitivity analysis to evaluate the consistency of the LLM’s discovered categories. We re-ran the entire categorization pipeline across three independent trials using different random seeds. Qualitative analysis of the inferred categories indicates that the ``top-N'' discovered categories, accounting for over 90\% of the total analyzed sequences, were inferred with remarkable consistency across all runs. For instance, core categories such as ``Reviewing comments (collaboration)'' and ``Content transfer within or between documents'' appeared in every trial. The variance in these high-frequency categories was limited strictly to lexical changes in naming; for example, one trial labeled a cluster as ``Active editing of content'' while another named it ``Editing content actively,'' yet both mapped to the same underlying distribution of low-level user action patterns as verified by tf-idf scores of the action sequences associated with these categories. In contrast, the ``long-tail'' activities representing less than 10\% of the dataset exhibited higher instability; for example, ``Document navigation'' was not consistently isolated as a standalone category across all runs, often being absorbed into broader clusters. While future work could involve extensive human evaluation to further validate these semantic boundaries, the qualitative results observed across our experiments—particularly the browser inference tasks in Table \ref{tab:mind2web-qual-example} and Table \ref{tab:mind2web-qual-examples}—are highly compelling. The precision with which the framework translates low-level user action sequences into human-readable intent builds significant confidence in the stability and quality of these inferred categories for functional product telemetry, where the goal is to understand the most common user journeys.

\begin{table*}
\small{
\begin{tabular}{p{2cm}|p{13cm}}
\toprule
\textbf{Prompt Type} & \textbf{Content} \\
\toprule
System Prompt & You are an expert at analyzing web interaction patterns and translating technical UI action logs into clear, human-readable descriptions of user behavior and intent. \\
\midrule
User Prompt & You are analyzing web interaction telemetry data from user sessions. Your task is to provide a detailed natural language description of what the user is doing on a website based on the sequence of UI actions they performed. \newline
\textbf{Action sequence (in order):} \{action\_text\} \newline
Each action follows the format: [element\_type] element\_description $\rightarrow$ ACTION\_TYPE: optional\_value \newline
\textbf{Instructions:}
\begin{itemize}
    \item Provide a detailed, step-by-step description of what the user is doing
    \item Focus on the user's workflow and intent behind the actions
    \item Interpret technical UI elements into plain language
    \item Pay attention to the sequence and progression of actions
    \item Describe what the user is trying to accomplish through these interactions
    \item Include details about form filling, navigation, searches, selections, etc.
\end{itemize}
Provide a comprehensive description of the user's web interaction workflow. \\
\bottomrule
\end{tabular}
}
\vspace{1mm}
\caption{WorkflowView Layer 1 prompt for browser task inference; obtaining natural language descriptions from action sequences.}
\label{apptab:layer1-mind2web}
\end{table*}

\begin{table*}[ht]
\small{
\centering
\begin{tabular}{p{2cm}|p{13cm}}
\toprule
\textbf{Prompt Type} & \textbf{Content} \\
\midrule
System Prompt &
You are an expert at distilling detailed user workflow descriptions into concise, natural task descriptions that capture the user's primary intent. \\
\midrule
User Prompt &
Based on the detailed workflow analysis below, generate a concise task description that captures what the user is trying to accomplish. The task description should be similar to how a user would naturally describe their goal when using a website. \newline

\textbf{Detailed workflow analysis:} \{detailed\_description\} \newline

\textbf{Instructions:}
\begin{itemize}
    \item Generate ONLY the task description, nothing else
    \item Make it concise and action-oriented
    \item Focus on the end goal, not the individual steps
    \item Use natural language that describes the user's intent
    \item Do not mention the website name, domain, or technical details
    \item Format it as a simple sentence or phrase describing the task
    \item Include the necessary details that are required to successfully complete the task
\end{itemize} 

\textbf{Task description:}\\
\bottomrule
\end{tabular}
}
\vspace{1mm}
\caption{WorkflowView Layer 2 Prompt for browser task inference; obtaining succinct summary of user intent.}
\label{apptab:layer2-mind2web}
\end{table*}

\begin{table*}[ht]
\small{
\centering
\begin{tabular}{p{2cm} | p{13cm}}
\toprule
\textbf{Prompt Type} & \textbf{Content} \\
\midrule
System Prompt & You are an expert at analyzing online learning behavior patterns and translating technical activity logs into clear, human-readable descriptions of student engagement and learning patterns. \\
\midrule
User Prompt & 
You are analyzing MOOC (Massive Open Online Course) learning behavior data. Your task is to provide a detailed natural language description of a student's learning activities and engagement patterns based on their chronological sequence of actions.

\textbf{Time Window Analyzed:} \{hours\_start\_before\_last\} to \{hours\_end\_before\_last\} hours before last action

\textbf{TARGET STUDENT'S ACTIVITY SEQUENCE:} \{action\_text\}

\textbf{Instructions:}
\begin{itemize}
\item Provide a detailed, step-by-step description of what the student is doing
\item Focus on the student's workflow and intent behind the actions
\item Interpret technical UI elements into plain language
\item Consider the context of students interacting with MOOC
\item Pay attention to the sequence and progression of actions, their frequency, and timestamps to note regularity / sparsity / consistency / inconsistency
\end{itemize}

Provide a comprehensive and balanced description of the student's MOOC learning behavior and engagement patterns.
\\
\bottomrule
\end{tabular}
}
\vspace{1mm}
\caption{WorkflowView Layer for MOOC student dropout prediction; obtaining natural language descriptions.}
\label{apptab:layer1-mooc}
\end{table*}

\begin{table*}[ht]
\small{
\centering
\begin{tabular}{p{2cm} | p{13cm}}
\toprule
\textbf{Prompt Type} & \textbf{Content} \\
\midrule
System Prompt & You are an expert at summarizing student learning behavior patterns into concise, actionable insights about student engagement in online courses. \\
\midrule
User Prompt &
Based on the detailed learning behavior analysis below, generate a concise summary that captures the key patterns of student engagement and learning behavior in this MOOC course. \newline
\textbf{Time Window:} \{hours\_start\_before\_last\} to \{hours\_end\_before\_last\} hours before last action \newline
\textbf{TARGET STUDENT'S DETAILED ANALYSIS:} \{detailed\_description\} \newline
\textbf{Instructions:}
\begin{itemize}
    \item Generate a concise summary (2--3 sentences) of the student's engagement patterns
    \item Focus on key behavioral indicators and learning activity patterns
    \item Highlight temporal patterns and engagement levels
    \item Concisely surface insights that may help infer if the student will dropout of the course of continue based on their engagement so far
\end{itemize}
\textbf{Engagement Summary:} \\
\bottomrule
\end{tabular}
}
\vspace{1mm}
\caption{WorkflowView Layer 2 for MOOC student dropout prediction; obtaining succinct activity summary.}
\label{apptab:layer-mooc-summary}
\end{table*}

\begin{table*}
\centering
\begin{tabular} {l | p{0.65\textwidth}}

\toprule
\textbf{\small 1. Action Sequence} & \tiny{[svg] → CLICK, [link] Your lists → CLICK, [link] Create a list → CLICK, [svg] → CLICK, [span] Walgreens → CLICK, [textbox] Add a title (Required) → TYPE: Walgreens, [img] → CLICK, [button] Next → CLICK, [link] Personal Care → CLICK, [svg] → CLICK, [img] → CLICK, [span] Add to list → CLICK, [checkbox] Walgreens New → CLICK, [button] Done → CLICK, [path] → CLICK, [path] → CLICK, [path] → CLICK, [svg] → CLICK, [img] → CLICK, [span] Add to list → CLICK, [checkbox] Walgreens New → CLICK, [button] Done → CLICK, [path] → CLICK, [link] View More → CLICK, [img] → CLICK, [span] Add to list → CLICK, [checkbox] Walgreens New → CLICK, [button] Done → CLICK, [button] Back → CLICK, [path] → CLICK, [link] Shower Essentials → CLICK, [img] → CLICK, [span] Add to list → CLICK, [checkbox] Walgreens New → CLICK, [button] Done → CLICK, [button] Back → CLICK, [link] Lists → CLICK}\\
\midrule
\textbf{\small Generated Task Description} & \small{Create a Walgreens shopping list and add personal care and shower essentials items.}\\
\midrule
\textbf{\small Ground-truth Task Description} & \small{Create a new list and add four items from the personal care category at Walgreens.} \\
\bottomrule

\toprule
\textbf{\small 2. Action Sequence} & \tiny{[link] SEARCH CARS → CLICK, [button] Distance \& Shipping → CLICK, [button] Change Location → CLICK, [textbox] Enter ZIP or State → CLICK, [textbox] Enter ZIP or State → TYPE: 07470, [button] set store → CLICK, [button] Distance \& Shipping → CLICK, [button] Nationwide → CLICK, [menuitem] 100 miles → CLICK, [input] → CLICK, [button] Back to all categories → CLICK, [button] Make → CLICK, [listitem] Honda → CLICK, [button] Back to all categories → CLICK, [button] Model → CLICK, [listitem] Civic → CLICK, [button] Back to all categories → CLICK, [button] Year → CLICK, [button] 2010 → CLICK, [menuitem] 2017 → CLICK, [button] 2023 → CLICK, [menuitem] 2017 → CLICK, [button] Back to all categories → CLICK, [button] Features → CLICK, [listitem] Sunroof(s) → CLICK, [button] Back to all categories → CLICK, [button] Exterior Color → CLICK, [listitem] Black → CLICK, [button] Back to all categories → CLICK, [button] Sort by → CLICK, [label] Lowest price → CLICK}\\
\midrule
\textbf{\small Generated Task Description} & \small{Find a black 2017 Honda Civic with a sunroof within 100 miles of Wayne, NJ, sorted by lowest price.}\\
\midrule
\textbf{\small Ground-truth Task Description} & \small{Find the cheapest 2017 Honda Civic within 100 miles of 07470 which has a sun roof and black exterior color while excluding out of market vehicles} \\
\bottomrule

\toprule
\textbf{\small 3. Action Sequence} & \tiny{[textbox] DESTINATION → TYPE: NIAGRA FALLS, [option] Niagara Falls, ON, Canada → CLICK, [button] Date selection → CLICK, [span] 17 → CLICK, [span] 20 → CLICK, [button] Done → CLICK, [button] 1 Room, 1 Guest → CLICK, [button] Add Rooms → CLICK, [button] Add Adults → CLICK, [button] Add Adults → CLICK, [button] Add Children → CLICK, [button] Add Age → CLICK, [button] Add Less than 1 → CLICK, [button] Add 1 → CLICK, [button] Add 2 → CLICK, [button] Find Hotels → CLICK, [checkbox] Show rates with taxes and all fees → CLICK, [checkbox] Show available hotels only → CLICK, [link] Free breakfast → CLICK, [link] Pool → CLICK, [generic] Price → CLICK, [label] 100 - 200 USD → CLICK, [button] APPLY → CLICK, [combobox] Sort by → CLICK, [option] Price → CLICK, [link] VIEW RATES → CLICK, [button] Member Rate Prepay Non-refundable → CLICK, [label] Accept cancellation → CLICK, [button] CONTINUE → CLICK}\\
\midrule
\textbf{\small Generated Task Description} & \small{Book a hotel in Niagara Falls, ON, for three adults and three children from April 17 to April 20 with free breakfast and a pool, within a \$100-\$200 budget after including taxes and fees.} \\
\midrule
\textbf{\small Ground-truth Task Description} & \small{Find two rooms in a cheapest hotel in Niagra Falls for three adults and one three year old kid from May 17 to May 20, view only available hotels within 100 to 200 dollar range with taxes and fees, and choose the cheapest hotel that offers free breakfast and a pool.} \\
\bottomrule

\toprule
\textbf{\small 4. Action Sequence} & \tiny{[link] Shop → CLICK, [img] Sports car icon → CLICK, [button] Sort by → CLICK, [label] Lowest price → CLICK, [button] Back to all categories → CLICK, [button] Fuel Type → CLICK, [listitem] Gas → CLICK, [button] Back to all categories → CLICK, [button] Year → CLICK, [button] 2010 → CLICK, [menuitem] 2018 → CLICK, [button] 2023 → CLICK, [menuitem] 2022 → CLICK, [button] Back to all categories → CLICK, [button] Exterior Color → CLICK, [listitem] Gray → CLICK, [button] Back to all categories → CLICK, [button] Transmission → CLICK, [span] Automatic → CLICK, [button] Back to all categories → CLICK, [heading] Distance \& Shipping → CLICK, [button] \$99 Or Less → CLICK, [menuitem] Free to home or store → CLICK, [button] Back to all categories → CLICK, [switch] COMPARE → CLICK, [path] → CLICK, [button] Add to Compare → CLICK, [button] Go to Compare → CLICK, [button] COMPARE PHOTOS → CLICK}\\
\midrule
\textbf{\small Generated Task Description} & \small{Find and compare affordable gray automatic sports cars that run on gas from recent years (2018-2022) with no shipping costs.} \\
\midrule
\textbf{\small Ground-truth Task Description} & \small{Search for an automatic grey sports car with the lowest price, gas fuel and free shipping manufactured between 2018 to 2022, compare the top two results and compare photos.} \\
\bottomrule

\toprule

\textbf{\small 5. Action Sequence} & \tiny{[button] hotels → CLICK, [div] Destination or property → TYPE: jakarta, [hp-input-button] Destination or property → TYPE: jakarta, [div] Jakarta → CLICK, [div] Choose date → CLICK, [div] Jun → CLICK, [generic] 1 → CLICK, [generic] 4 → CLICK, [button] Search → CLICK, [button] Yes, I agree → CLICK, [span] Lowest price → CLICK, [button] Choose room → CLICK, [button] Book now → CLICK, [textbox] First name → TYPE: Joe, [textbox] Surname → TYPE: Bloggs, [textbox] Email address → TYPE: buckeye.foobar@gmail.com, [textbox] Confirm email address → TYPE: buckeye.foobar@gmail.com, [input] → TYPE: 1111111111111111, [textbox] Address 1 → TYPE: the home of joe bloggs, [textbox] City → TYPE: new york, [textbox] Postcode/ZIP code → TYPE: 10001, [combobox] State → TYPE: new york, [textbox] Card number → TYPE: 1234, [combobox] Card type → SELECT: MasterCard, [combobox] Month → SELECT: 01, [combobox] Year → SELECT: 2023, [textbox] CVV → TYPE: 123, [textbox] Cardholder's name → TYPE: joe bloggs, [svg] → CLICK}\\
\midrule
\textbf{\small Generated Task Description} & \small{Book a hotel in Jakarta from June 1 to June 4 at the lowest price, for Joe Bloggs with email buckeye.foobar@gmail.com and phone number 11111111111. The billing address is specified to be in New York, 10001.}  \\
\midrule
\textbf{\small Ground-truth Task Description} & \small{Book the cheapest available hotel for a three night stay from 1st June in Jakarta. The guest is named Joe Bloggs with the email address of buckeye.foobar@gmail.com and phone number of 11111111111. Billing address is in New York, zip code 10001.} \\
\bottomrule
\end{tabular}
\caption{\small Qualitative examples of task descriptions generated using WorkflowView from the action sequences, and the corresponding ground-truth task descriptions. }
\label{tab:mind2web-qual-examples}
\end{table*}
\end{document}